\pdfoutput=1

\documentclass[11pt]{article}
\usepackage{listings}
\usepackage[preprint]{acl}
\usepackage{caption}

\usepackage{comment}
\usepackage{verbatim}
\usepackage{times}
\usepackage{latexsym}
\usepackage{algorithm}
\usepackage{algorithmic}
\usepackage{subcaption}
\usepackage{graphicx}
\usepackage{makecell} 
\usepackage[T1]{fontenc}
\usepackage{amsmath}
\usepackage{amsfonts}
\usepackage[utf8]{inputenc}
\usepackage{tabularx}
\usepackage{soul}

\usepackage{microtype}

\usepackage{inconsolata}
\usepackage{booktabs}

\usepackage{multirow}

\usepackage{graphicx}

\AtEndPreamble{
    \usepackage[capitalize]{cleveref}
    \crefname{section}{Sec.}{Secs.}
    \Crefname{section}{Section}{Sections}
    \Crefname{table}{Table}{Tables}
    \crefname{table}{Tab.}{Tabs.}
}
\title{Can a Unimodal Language Agent Provide Preferences to Tune a Multimodal Vision-Language Model?}

\author{Sazia Tabasum Mim\thanks{\;Equal contribution.} \\
Georgia State University \\
\texttt{smim2@student.gsu.edu}
\And
Jack Morris\footnotemark[1]\\
Georgia State University \\
\texttt{jmorris116@gsu.edu}
\And
Manish Dhakal \\
Georgia State University \\
\texttt{mdhakal3@gsu.edu}
\AND
Yanming Xiu \\
Duke University \\
\texttt{yanming.xiu@duke.edu}
\And
Maria Gorlatova \\
Duke University \\
\texttt{maria.gorlatova@duke.edu}
\And
Yi Ding \\
Georgia State University \\
\texttt{yding@gsu.edu}
}

\begin{document}
\maketitle
\begin{abstract}

To explore a more scalable path for adding multimodal capabilities to existing LLMs, this paper addresses a fundamental question: Can a unimodal LLM, relying solely on text, reason about its own informational needs and provide effective feedback to optimize a multimodal model? To answer this, we propose a method that enables a language agent to give feedback to a vision-language model (VLM) to adapt text generation to the agent's preferences.  Our results from different experiments affirm this hypothesis, showing that LLM preference feedback significantly enhances VLM descriptions. Using our proposed method, we find that the VLM can generate multimodal scene descriptions to help the LLM better understand multimodal context, leading to improvements of maximum 13\% in absolute accuracy compared to the baseline multimodal approach. Furthermore, a human study validated our AI-driven feedback, showing a 64.6\% preference alignment rate between the LLM's choices and human judgments. Extensive experiments provide insights on how and why the method works and its limitations.

\end{abstract}

\section{Introduction}
\label{sec:intro}
Large Language Models (LLMs) or Language Agents have emerged as powerful tools for processing and generating textual data. They have significantly advanced natural language processing by achieving near human-level performance across a wide range of text-centric tasks, including text classification, reasoning, and open-domain question answering, content generation, and others ~\cite{guo2025deepseek,brown2020language,radford2018improving, dam2024complete,qian2025vc,wang2025end,jansen2025leveraging,huang2025building}. An active area of research is to make these agents multimodal~\cite{li2023blip,achiam2023gpt,liu2024improved} to support a wider range of human-AI interactive tasks. However, a key limitation persists: these models typically require training from scratch, or extensive retraining of pre-existing LLMs \cite{bender2021dangers}, which is costly \cite{xia2024understanding} and prohibitive to train for many. 

To address this challenge, existing multimodal pipelines often convert visual or acoustic inputs into text so that LLMs can reason over them. Prior works such as CLIP \cite{radford2021learning}, TextMI \cite{kamrul2023textmi}, and Hyper-LLaVA \cite{zhang2024hyperllava} leverages the reasoning capabilities of unimodal LLMs by converting multimodal inputs into text descriptions. By adopting this method, we would only need to train lightweight adaptors that describe multimodal features with text, enabling a unimodal LLM to reason over them without requiring costly multimodal retraining. A key advantage of this method is the interpretability of the text descriptions, which serve as a "latent" modality, offering insights into the LLM's decision-making process.

However, a significant challenge arises: unoptimized VLM descriptions often introduce noise, focusing on irrelevant details or failing to capture subtle contextual cues essential for tasks like sarcasm detection. Our experiments show that naively incorporating such descriptions can degrade LLM performance compared to text-only baselines ( \cref{tab:vlm_accuracy}). To overcome this, we introduce a novel framework where a unimodal LLM provides preference feedback to optimize VLM-generated descriptions. Using Direct Preference Optimization (DPO) \cite{rafailov2023direct}, we fine-tune the VLM to produce descriptions tailored to the LLM's needs for downstream reasoning tasks. Our DPO-based tuning focuses on generating a single, preference-aligned description that reflects the LLM’s feedback signal. This approach ensures that the VLM generates concise, relevant, and task-specific descriptions, avoiding the pitfalls of generic or noisy outputs.

We evaluate our method on two multimodal social reasoning datasets: MUStARD \cite{castro2019towards} for sarcasm detection and UR Funny \cite{hasan2019ur} for humor detection. These datasets serve as challenging testbeds due to their reliance on subtle, non-verbal cues \cite{liu2022towards}. To demonstrate generalizability, we test our framework across multiple LLM architectures and demonstrate consistent performance improvements. For instance, on the MUStARD dataset, our method achieves a sarcasm detection accuracy of 66.91\% (\cref{tab:performance_comparison}), compared to 51.80\% for the zero-shot VLM baseline. Similar gains are observed on UR Funny, highlighting the robustness of our approach across tasks and models. 

\noindent The main contributions of this study are as follows:
\begin{itemize}
    \item We propose the first framework that allows a unimodal LLM to provide preference feedback for fine-tuning a multimodal VLM.
    
    \item We explore different feedback settings, highlighting how textual agents can guide visual optimization without direct access to multimodal data.
    
    \item We empirically demonstrate that preference tuning improves both reasoning performance and visual faithfulness.

\end{itemize}

\section{Related Work}

\subsection{Reinforcement Learning from AI Feedback (RLAIF)}
Reinforcement Learning from Human Feedback (RLHF) is a standard method for aligning LLMs with human preferences using annotated pairs to train a reward model \citep{christiano2017deep,stiennon2020learning,ouyang2022training}. \citet{bai2022constitutional} and \citet{lee2023rlaif} introduced Reinforcement Learning from AI Feedback (RLAIF), in which off‐the‐shelf LLMs replace costly human labelers to generate preference labels. Empirical results show that RLAIF matches or exceeds RLHF on summarization, helpfulness and harmlessness benchmarks reducing annotation cost and improving scalability \citep{lee2023rlaif,gilardi2023chatgpt}. Unlike RLAIF, which leverages AI feedback to optimize model outputs for purely language‐generation tasks, our work uses LLM‐driven preferences to shape the inputs to an LLM‐based classifier in a multimodal setting. We use the resulting preference pairs, not to fine‐tune the LLM’s generation policy, but to refine the VLM so that it produces better descriptions.  

\subsection{LLM-Based Multimodal Reasoning}
LLMs have shown remarkable capabilities in understanding and generating text, prompting research into their potential for multimodal reasoning. \citet{yang2022empirical} explored LLMs for multimodal tasks by converting images into textual captions using models like CLIP \citet{radford2021learning}, then feeding these captions into LLMs for tasks such as visual question answering. Similarly, \citet{zhang2024hyperllava} used VLMs to generate textual descriptions of images, which were then processed by LLMs for reasoning tasks.  \citet{kamrul2023textmi} propose TextMI, an innovative framework designed to convert acoustic and visual information into textual descriptions, allowing these cues to be effectively processed by text-based models like BERT. These studies focus on static image-based tasks and do not address dynamic video contexts or complex social cues like sarcasm. Our approach extends this line of work by applying LLM-based reasoning to task requiring temporal and contextual understanding.

\subsection{Preference Optimization in Model Alignment}
Direct Preference Optimization (DPO) has emerged as a powerful technique for aligning models with human or AI-generated preferences, bypassing the need for complex reward modeling \citep{rafailov2023direct}. In the context of VLMs, \citet{zhou2024aligning} employed DPO to fine-tune vision-language models by generating preference data using GPT-4V, targeting hallucination reduction. \citet{zhang2024direct} applied DPO to video multimodal models for optimizing instruction-following capabilities. While these works leverage DPO for VLM alignment, they focus on direct multimodal input processing rather than text-based abstractions.Our study innovates by integrating DPO into an LLM feedback loop in a novel way: we refine the VLM's generated textual descriptions that serve as inputs to a downstream LLM-based tasks. This feedback-driven approach to structuring textual inputs enables the LLM to better interpret multimodal contexts and improves its performance.

Unlike prior studies that either convert multimodal signals into text for downstream reasoning without improving the underlying visual model, or apply preference alignment only to language generation, our work introduces a fundamentally different perspective. We show that a purely unimodal language agent, without direct access to visual inputs, can still provide effective preference feedback to tune a multimodal VLM itself, rather than merely using its outputs. To our knowledge, our approach is the first to systematically demonstrate that a unimodal LLM can guide a VLM toward producing more task-relevant and visually aligned descriptions

\section{Methodology}
\label{sec:methodology}

\begin{figure*}[htbp]
    \centerline{
    \includegraphics[width=1\textwidth]{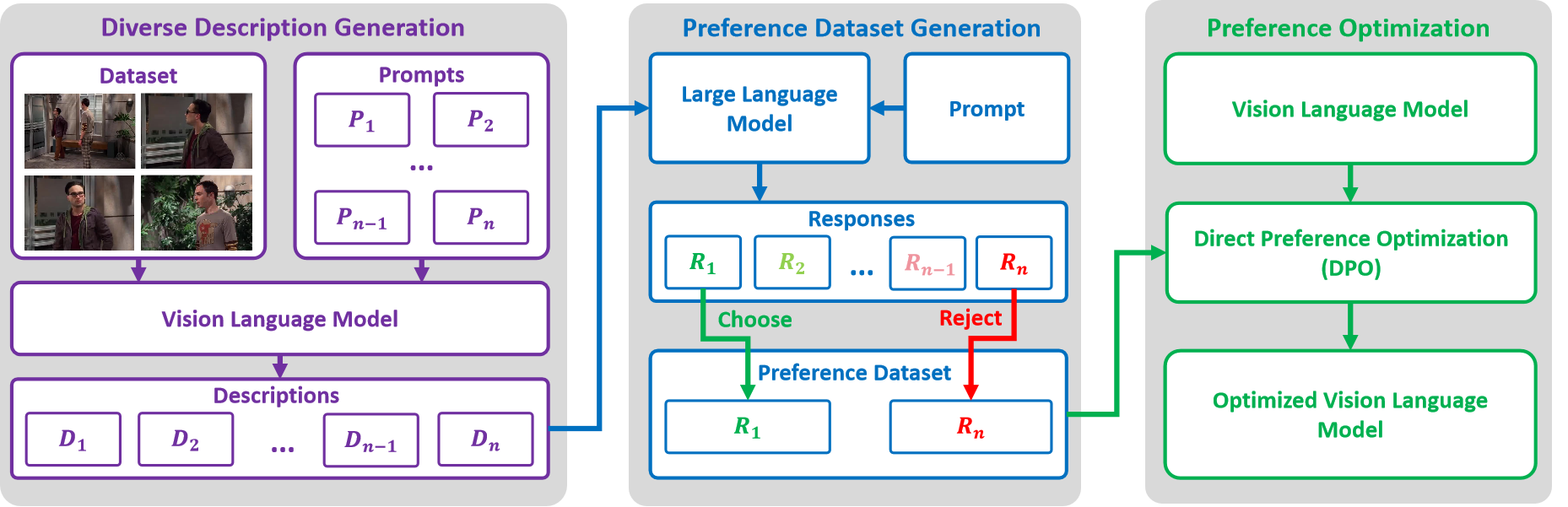}}
    \caption{An overview of the training pipeline. (a) First, a VLM generates diverse descriptions with a set of prompts; Second, (b) descriptions are ranked using an LLM Agent to generate preference dataset; Lastly, (c) generated preference dataset is used for DPO-based optimization to fine tune the VLM. The agent model stays fixed.}
    \label{fig:framework}
\end{figure*}

Our methodology for preference tuning follows a similar pipeline as RLHF \cite{ouyang2022training} and other preference tuning methods \cite{rafailov2023direct}. There are two important differences: first, we do not use any demonstration (i.e. agent does not generate text of what is preferred since it does not have access to the visual modality); and second, we only use a unimodal language agent for preference feedback. We are interested in testing if a unimodal language agent, which has no direct access to the visual modality, can effectively guide a multimodal model. This setup allows us to investigate our core research question while also presenting a cost-effective, and computationally efficient alternative to traditional alignment techniques that require extensive human annotation or the training of massive, end-to-end multimodal systems. 

To answer whether a unimodal language agent can provide feedback to a multimodal vision language model (VLM) we setup an experiment with the following methodology (visualized in \cref{fig:framework}):

\textbf{Generating Captions with a VLM:} We first ask the VLM (e.g. Llava) to generate multiple language descriptions of a video. 

\textbf{Generating Agent Preferences:} We then ask a language agent (e.g. DeepSeek R1) to rank generated text descriptions using a few prompt variations.

\textbf{Preference Optimization:} Using the preferences provided by the language agent, we optimize the VLM using preference optimization (e.g. DPO).

\subsection{Generating Captions with a VLM}
The vision language model (VLM) is a model which maps a short video segment into a caption i.e.:
\[
\pi_\theta: x_v \in \mathbb{R}^{v_w \times v_h \times v_t} \to x_t \in \mathbb{R}^{l \times h},
\]
where the input video segment is a sequence of $v_t$ images with width and height of $v_w$ and $v_h$. The output caption ($x_t$) is of length $l$ with embedding dimension $h$. The images are sampled uniformly (8 frames per video segment frames). 

To increase the diversity of the generated prompts for the video content, we design five distinct prompts. We target different aspects of the video including general content, emotional cues, and facial expressions.  For each video, we apply a conversational template combining both the video content and a set of crafted prompts. We run the model across these multiple prompts to obtain a set of five unique description outputs, which are then stored for downstream tasks. For example, we ask the VLM to \texttt{"Describe what is happening in this video in detail."} or \texttt{"Describe the facial expressions in the video that might indicate contrasting emotions. Keep the description brief"}. See \cref{sec:vlm-prompts} for additional prompts that were used. We do not use any sampling strategies (such as varying temperature or top-k) as we noticed that these contributed minimally to differences in generated captions. Instead we only rely on the distinct prompt formulations.

\subsection{Generating Agent Preferences}\label{pref}
We are interested in assessing 1) whether agents are capable of providing preferences to tune another model, and 2) how capable they are. To do so, we propose three conditions for evaluating the preferences provided by agents. First, agents provide preferences \textit{with} knowledge of what is the ground-truth class of the dataset. Second, agents provide preferences on VLM outputs \textit{without} knowing what is the true class of the dataset. And lastly, we provide in-context examples of preferences without knowledge of ground truth.

In all conditions, the language agent remains strictly unimodal. It does not have access to the visual information and relies solely on text inputs. The conditions differ only in the information provided to the agent alongside the VLM's generated descriptions. The ground-truth (GT) label is used only in the `With GT' condition (Sec \ref{with_gt}) as an explicit, text-based guide for preference generation. In our primary `Without GT' condition (Sec \ref{ranking no gt}), the unimodal agent has access to neither the true labels nor the visuals, allowing us to purely evaluate how such an agent can provide effective feedback.

\subsubsection{Agent preferences with ground truth knowledge (\texttt{With GT})}\label{with_gt}

In this approach, we incorporate the known ground-truth label to guide the LLM’s evaluation. For each video, we provide the five generated descriptions alongside text-based conversation and the ground-truth label. The LLM is then prompted to rank how helpful each description is in predicting the known label
(see \cref{prompt2} for detailed prompt).

\subsubsection{Agent preferences without ground truth knowledge (\texttt{Without GT})} \label{ranking no gt}

Then we aim to evaluate whether an LLM can independently assess the likelihood of sarcasm in a scene using only visual text description and conversational context, without providing any ground-truth labels. For each video, we supply the LLM with the generated video description paired with previous context and original utterance from the scene.

In this method, the LLM is asked to predict the likelihood of ground-truth on a scale from 1 to 10. Formally, given input $q = (d, u, c)$, where $d$ is the description, $u$ is the utterance, and $c$ is the previous context of the given video. The LLM outputs a scalar score $s \in [1, 10]$ estimating the nature of the scene (e.g. sarcastic for MUStARD dataset). For example,  1 means that the LLM strongly believes $u$ \emph{is not} sarcastic and 10 means strongly believes $u$ \emph{is} sarcastic (see \cref{prompt1} for detailed prompt). According to the algorithm \ref{alg:ranking}, we rank descriptions based on ground truth and score, $s$.

\subsubsection{Agent preferences with in-context examples (\texttt{With ICL})}

In this approach, we adopt an in-context learning (ICL) approach by providing the language model with two illustrative examples to demonstrate the ranking process for video descriptions. For MUStARD dataset, the first example describes a scene containing sarcasm, which contains a score of 10 to indicate strong sarcastic content. The second example describes a scene without any sarcasm, with a score of 1 to represent no sarcasm. Then, we prompt the LLM to predict the likelihood of sarcasm as described in Section \ref{ranking no gt}.

For our in-context examples, we selected two of the model’s raw outputs from Section \ref{ranking no gt}. One is of score 10 for a sarcastic scene and another is scored 1 for a non-sarcastic scene. The full prompt can be found in Appendix \ref{prompt4}

\subsubsection{Sorting preferences}
We provide a simple rule to determine the order of the preferences (found in \cref{alg:ranking}). These rules are applied to the \texttt{Without GT}  and \texttt{With ICL} conditions. The reason we do this is to evaluate how effective agents are at incorporating the ground truth. The ranking provides a (weak) set of rules that restricts the preferences to align with the ground truth of the dataset manually.

\begin{figure*}[tbp]
    \centerline{
    \includegraphics[width=1\textwidth]{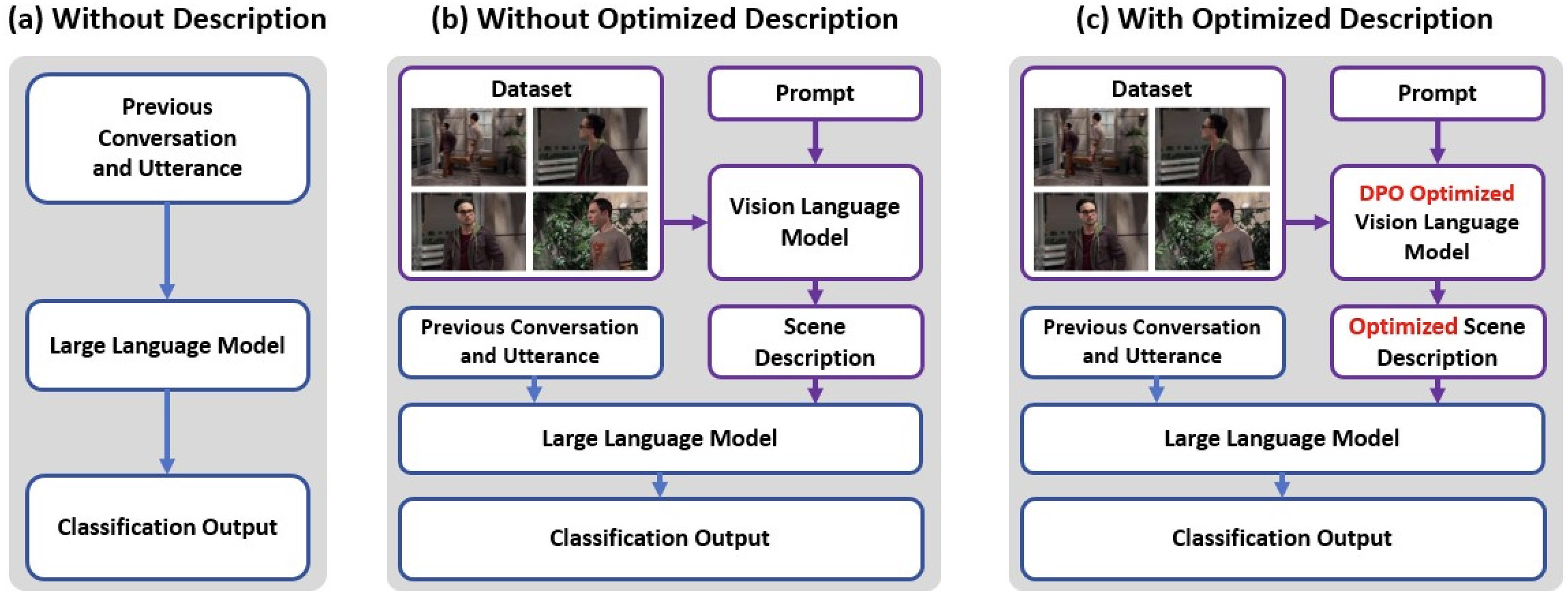}}
    \caption{Overview of the evaluation process. In the unimodal evaluation pipeline shown in (a), only the conversation context and the final statement is given to the Agent to determine the label. (b) demonstrates how multimodal information is incorporated during the evaluation process. The VLM acts like a feature extractor by converting visual features to text descriptions. This is fed to the LLM. (c) shows how the preference tuned VLM (Optimized Vision Language Model) is incorporated into the pipeline.}
    \label{fig:eval-process}
\end{figure*}

\begin{algorithm}[t]
\small
\caption{Select Preferred Description}
\begin{algorithmic}[1]
\label{alg:ranking}

\STATE Sort $scores$ list in ascending order by score

\IF{ground\_truth\_sarcasm is True}
    \STATE preferred $\leftarrow$ description with highest score (last in $scores$ list)
    \STATE dispreferred $\leftarrow$ description with lowest score (first in $scores$ list)
\ELSE
    \STATE preferred $\leftarrow$ description with lowest score (first in $scores$ list)
    \STATE dispreferred $\leftarrow$ description with highest score (last in $scores$ list)
\ENDIF
\end{algorithmic}
\end{algorithm}

\subsection{Preference Optimization}
Direct Preference Optimization (DPO) \cite{rafailov2023direct} is used to align VLM outputs to the preferences of the language agent. To summarize, DPO leverages a reparameterization of the reward function under the Bradley-Terry model, showing that the optimal policy satisfying human preferences can be obtained by maximizing a simple binary cross-entropy objective:

\begin{small}
\begin{align}
\mathcal{L}_{\mathrm{DPO}}(\pi_\theta; \pi_{\mathrm{ref}})
= \nonumber \\ - \mathbb{E}_{(x, y_w, y_l) \sim \mathcal{D}} \Big[\log \sigma \Big(
&\beta \log \frac{\pi_\theta(y_w|x) \cdot \pi_{\mathrm{ref}}(y_l|x)}{\pi_{\mathrm{ref}}(y_w|x) \cdot \pi_\theta(y_l|x)}
 \Big) \Big] ,
\end{align}
\end{small}

\noindent where $y_w$ and $y_l$ denote the preferred and dispreferred responses, respectively, $\pi_{\mathrm{ref}}$ is the reference policy (often the supervised fine-tuned model), and $\beta$ controls the KL-divergence regularization strength. This formulation allows DPO to directly adjust the model’s log-probabilities to satisfy human preferences without maintaining a learned reward critic.

In our case, $\pi_\theta$ is the VLM we want to fine-tune and $\pi_{\mathrm{ref}}$ is a reference model, or an unmodified version of $\pi_\theta$. $D$ is the dataset of preferences generated in the section \ref{pref}. $x$ is a sample prompt which is "Describe the video in details". $y_w$ and $y_l$ are the LLM's preferred and disprefered response to the prompt $x$. Using $\beta$ we control the amount of divergence from the reference model $\pi_{\mathrm{ref}}$.

\subsection{Agent Evaluation Process}

We wish to see if the agent is properly accounting for the visual context to make a decision about a multimodal property. Therefore, both the textual description of the visual context as well as the original text-based caption should be available to the agent for it to make its decision. Detailed prompt is provided in \cref{prompt4}.

 In this way, we can evaluate how well the agent can integrate the visual descriptive context with the utterance. We present the conditions of evaluation in \cref{fig:eval-process}. Once the LLM provides its output, we assess our model’s performance using standard classification metrics (precision, recall, accuracy) following the tasks of the dataset.

\section{Experimental Setup}
\label{sec:experimental-setup}

\subsection{Dataset}

We validate our framework on two distinct social multimodal datasets. For sarcasm detection, we use the MUStARD \citet{castro2019towards} dataset, a balanced collection of 690 video clips from popular TV shows, and follow split of 551 clips for training and 139 for testing. To test for generalizability on a different task, we use the UR-FUNNY \cite{hasan2019ur} dataset for humor detection. This dataset contains 16,514 balanced instances of humorous and non-humorous punchlines sourced from 1,866 TED Talks. For our experiments with UR-FUNNY, we adhere to the standard speaker-independent splits provided by the authors.

These datasets are particularly suitable for our study because they emphasize context-dependent multimodal reasoning rather than low-level perception. In both MUStARD and UR-FUNNY, the interpretation of an utterance depends on subtle visual and situational cues, such as facial expressions, gestures, or tone that can change its meaning depending on the context. This property makes them ideal testbeds for evaluating whether a unimodal language agent, operating purely on text, can provide effective feedback to improve a vision-language model’s descriptive capabilities.

\subsection{Vision Language \& Agent Model}
In our experiment, we fine tune the LLaVA-Video model, a large-scale video-language model developed to advance instruction-following tasks in multimodal settings \cite{zhang2024video}. The model architecture uses a specialized video representation technique called SlowFast, allowing the system to process up to three times more frames than standard methods within GPU memory constraints. To generate video descriptions during evaluation we use the following prompt for all methods:

\begin{lstlisting}
Describe the speaker's nonverbal cues, the context, and any mismatches between them.
\end{lstlisting}
To validate the generalizability of our approach, we evaluated our framework with six different open-source LLM agents of varying architectures and sizes.

\subsection{Training Details:}
We conducted our experiments on a system equipped with six NVIDIA GeForce RTX 4090 GPUs (24GB VRAM each) using CUDA 12.6. We used CUDA for GPU acceleration with fallback to CPU. To fine-tune our model, hyperparameters that impacted computational constraints were chosen to balance algorithmic performance and computational feasibility. 

We fine-tuned the VLM for 5 epochs using a batch size of 1, a learning rate of $1 \times 10^{-5}$, and the generic prompt \textit{"Describe the video in detail"}. We fine-tuned the VLM using Low-Rank Adaptation (LoRA) with a rank of \(r=4\), a scaling factor of \(\alpha=16\), and a dropout probability of 0.1. LoRA was applied to the query, key, and value projection layers (\texttt{q\_proj}, \texttt{k\_proj}, and \texttt{v\_proj}).

To ensure consistency and mitigate randomness in the evaluation process, we set the temperature to 0.00 for language agents.
This configuration maintains reliability in evaluation and ensure meaningful comparisons across different samples.

\section{Result Analysis}

\subsection{Main Results}
\begin{table*}[t]
\centering
\small
\begin{tabular}{@{}l|ccc|ccc@{}} 
\toprule
\textbf{Model} & \multicolumn{3}{c}{\textbf{MUStARD dataset}} & \multicolumn{3}{c}{\textbf{UR-FUNNY dataset}} \\
\cmidrule(lr){2-4} \cmidrule(lr){5-7}
 & \begin{tabular}[c]{@{}c@{}}Baseline \\ Multimodal\end{tabular} & \begin{tabular}[c]{@{}c@{}}Utterance \\ only\end{tabular} & \begin{tabular}[c]{@{}c@{}}Preference tuned \\ Multimodal (ours)\end{tabular} & \begin{tabular}[c]{@{}c@{}}Baseline \\ Multimodal\end{tabular} & \begin{tabular}[c]{@{}c@{}}Utterance \\ only\end{tabular} & \begin{tabular}[c]{@{}c@{}}Preference tuned \\ Multimodal (ours)\end{tabular} \\
\midrule
Deepseek-r1-7b  & 52.9 & 61.2 & \textbf{66.9} & 55.6 & \textbf{57.8} & 56.5 \\
Qwen-7b         & 53.3 & 51.8 & \textbf{61.2} & 48.0 & 48.3 & \textbf{49.8} \\
Vicuna-30b      & 48.3 & \textbf{56.1} & 52.7 & 49.4 & 49.9 & \textbf{50.7} \\
Lama2-7b        & 50.8 & 53.5 & \textbf{56.5} & 49.0 & 49.5 & \textbf{53.5} \\
Llama3.3-70b    & 64.0 & \textbf{71.9} & 66.2 & 67.8 & \textbf{73.0} & 69.6 \\
Gemma2-2b       & 53.7 & 52.5 & \textbf{56.3} & 50.5 & 50.1 & \textbf{57.7} \\
\bottomrule
\end{tabular}%
\caption{Accuracy (\%) of Different agents on Sarcasm and Humor Detection. The table demonstrates that our proposed fine-tuning method consistently improves performance over the baseline multimodal (a combined input of both the speaker's utterance and the VLM-generated textual description) approach. For many agents, the fine-tuned descriptions help surpass the utterance only (only the transcribed text of the speaker's final spoken line) accuracy, while in other cases, the utterance-only performance remains the highest, even though the fine-tuned description is still a substantial improvement over the un-tuned one.}
\label{tab:vlm_accuracy}
\end{table*}

To assess the impact and generalizability of our proposed framework, we evaluated performance across six different agents on two distinct multimodal tasks: sarcasm detection and humor detection. The preference-tuned VLM, discussed in this section is trained using \texttt{Without GT}  preference dataset. The high-level accuracy results are summarized in \cref{tab:vlm_accuracy}.

The findings reveal a consistent and powerful trend. Our preference-tuning method provides a clear and robust improvement over the baseline. For all six agents on both datasets, the Preference tuned Multimodal condition significantly outperforms the Baseline Multimodal condition. This universally positive result demonstrates that our framework successfully aligns the VLM’s output with the agent’s needs, turning the visual description from an unreliable signal into a valuable one. Furthermore, for a majority of the agents, the performance of the preference-tuned model surpasses the strong Utterance only baseline, confirming the overall effectiveness of our framework. To validate our approach, we measured the zero-shot VLM performance (51.80\% on MUStARD) showing that our method outperforms the base VLM performance as well (see \cref{vl-eval}). For a more granular analysis, we provide a detailed breakdown of all performance metrics including Visual Only (only the textual description of the video generated by the VLM) conditions in \cref{tab1} in \cref{sec:other-results}. 

Further to understand why the preference-tuned descriptions lead to better downstream performance, we conducted a direct analysis of their semantic similarity to the source videos. The detailed experiment is provided at section \ref{what_changes_vlm}. 

\subsection{Impact of Different Modalities  \& Preference Datasets}
To understand the effect of different input modalities and preference data generation methods, we conducted a detailed analysis using our primary agent, \textit{Deepseek-r1}, on the \textit{MUStARD dataset}. We compare the agent's performance across utterance-only, visual-only and multimodal inputs. In multimodal input we include both the utterance and description of the scene. The utterance-only model reaches 61.2\% accuracy, while we see performance degradation with visual-only information (Tab. \ref{tab:performance_comparison}). Then we investigate how  descriptions generated by three different preference-tuned models (“Without GT,” “With GT,” and “With ICL”) help to alter the performance of the agent. With our training process using “Without GT” preference tuned model we get the highest accuracy (66.9\%)in the multimodal setting.

\begin{table}[t]
    \centering
    \small
    \begin{tabular}{l|ccc}
        \toprule
        \textbf{Model} & \textbf{Accuracy} & \textbf{Precision} & \textbf{Recall} \\
        \midrule
        \textbf{Baseline}   & & & \\
        \midrule
        Utterance Only & 61.2 &  62.0 &  64.0 \\
         Visual Only  & 48.9 &  52.0 & 19.0  \\
         Multimodal         & 52.9 &  58.0 &  40.0  \\ \midrule
        \textbf{Preference Tuned} & & & \\ \midrule
        \textbf{Visual Only} & & &  \\
        \quad With GT             & 47.48 &  49.0 & \ 40.0  \\
        \quad Without GT          & 50.36  & 71.0  & 7.0  \\
        \quad With ICL       & 51.80 &  65.0 &  15.0 \\
        \textbf{Multimodal}  & & & \\
        \quad With GT             & 58.7 &  58.0  & \  \textbf{75.0} \\
        \quad Without GT          & \textbf{66.9} &  70.0 &  65.0  \\
        \quad With ICL       & 65.1 &  \textbf{76.0}  & 50.0  \\
        \bottomrule
    \end{tabular}
    
    \caption{Performance comparison of baseline (pretrained VLM) compared to a few preference tuned conditions. While the visual-only modality did not outperform an utterance-only prediction, it improves over the descriptions obtained from an unoptimized VLM. Additionally, we see that there is better contextual information that the agent can make use of which led to the improvements in multimodal performance. \textbf{Note:} The preference-tuned models have to do with how the training dataset was obtained (see methods). \texttt{GT} information was not available at test time, nor during prompting or training of the VLM.}
    \label{tab:performance_comparison}

\end{table}

\subsection{Impact of the Temperature Parameter \texorpdfstring{($\beta$)}{(beta)} on DPO Fine-Tuning}

Here we highlight the effect of different $\beta$  values for our DPO fine tuning approach. Higher values of $\beta$ heavily penalize any deviation from the reference policy, whereas lower values encourage the model to diverge more to satisfy the reward signal.
To measure the impact of the DPO fine-tuning parameter, ($\beta$) with the preference dataset- ``Without GT'', we fixed the training prompt and inference prompt to \textit{"Describe the video in detail."} Then we ran the three experiments with $\beta$ set to 0.2, 0.5 and 0.8. As $\beta$ increased, the model's (deepseek-r1) accuracy steadily declined from 60.45\% to 53.33\% (\cref{tab:effect-of-different-beta-value} in the appendix). This indicates that stronger regularization toward the base policy yields more conservative but less adaptive behavior.

\subsection{Impact of Prompt Variations on performance}

To assess how different prompting strategies impact DPO fine-tuning and inference for a specific task, we experiment with three prompt variants: 1) Generic prompt ("\textit{Describe the video in detail.}"); 2)Task-specific prompt ("\textit{Describe the speaker’s nonverbal cues, the context, and any mismatches between them.}"\textit{)}; 3) No prompt. We use $\beta$ of 0.1 for these experiments.

\textbf{\textit{Effect of Prompt in DPO Fine-tuning:}} First, we assess how the training prompt affects our "Without GT" preference-tuned model by comparing three configurations: using a generic prompt, a task-specific prompt, and no prompt.

We used the same prompt for inference as used in the training process. We observe that training with the generic prompt yields the highest accuracy (60.45\%) (Tab. \ref{tab:effect-prompt-training}).

\begin{table}[t]
  \centering
  \small
  \begin{tabular}{l|ccc}
    \toprule
    \textbf{Training Prompt} & \textbf{Acc \%} & \textbf{Prec.} & \textbf{Recall} \\
    \midrule
    \makecell[l]{``Describe the speaker’s \\nonverbal cues, the context, \\and any mismatches \\ between them.''}
      & 58.33 & 60.0 & \textbf{63.0 }\\ \midrule
    ``Describe the video in detail.'' & 
    \textbf{60.45} & \textbf{63.0} & 55.0 \\ \midrule
    No Prompt    &        47.79       &    50.0   &    49.0      \\
    \bottomrule
  \end{tabular}
  \caption{Effect of Different Prompts on DPO Fine-Tuning. We choose the \texttt{"With GT"} preference tuned model to test the effect of different prompts in the DPO fine-tuning process. We notice that, with a generic prompt in the training process we can acheive higher accuracy.}
  \label{tab:effect-prompt-training}
\end{table}

\begin{table}[t]
    \centering
    \small
    
    \begin{tabular}{l|ccc}
        \toprule
        \textbf{Model} & \textbf{Accuracy} & \textbf{Precision} & \textbf{Recall} \\
        \midrule
        \textbf{Inference Prompt: P1}   & & & \\
        Base & 52.9 & 53.0 &53.0\\
         With GT  & 57.7 &  59.0 & 65.0  \\
         Without GT & 60.5 &  63.0 &  55.0 \\
         With ICL        & 60.0 &  64.0 &  51.0  \\ \midrule
        \textbf{Inference Prompt:P2}   & & & \\
        Base & 51.9 & 58.0 &40.0\\
         With GT  & 57.7 &  58.0 & \textbf{75.0}  \\
         Without GT & \textbf{66.9 }&  70.0 &  65.0 \\
         With ICL         & 65.1 &  \textbf{76.0} &  50.0  \\ 
        \bottomrule
    \end{tabular}
    \caption{Effect of different inference prompts in VLM. Here, prompt, P1="Describe the video in detail"; P2="Describe the speaker’s nonverbal cues, the context, and any mismatches between them". We observe that training with a generic prompt, we can achieve better performance with a task specific prompt during the inference time}
    \label{tab: effect-prompt-inference}
    
\end{table}

\textbf{\textit{Effect of Prompt During Inference:} }Here we evaluate inference‐time prompt effects on different preference tuned models while keeping the training prompt the same (\textit{"Describe the video in detail"}). Notably, when the model is trained with the generic prompt, switching to the task‐specific inference prompt significantly boosts performance (66.9\%). This demonstrate that a general training instruction combined with a focused, task‐specific inference prompt can yield superior results (Tab. \ref{tab: effect-prompt-inference}).

\subsection{What changed in the VLM outputs?} \label{what_changes_vlm}

To understand why the preference-tuned descriptions lead to better downstream performance, we analyzed their semantic and stylistic properties. First, we conducted a direct analysis of their semantic similarity to the source videos. Using the CLIP model (ViT-B/32), we measured the alignment between the video frames and the corresponding text descriptions from both the baseline VLM and our fine-tuned VLM. This analysis revealed that our fine-tuned descriptions are demonstrably more faithful to the videos. Across 139 videos from our test set, the fine-tuned (tuned ``Without GT'' on MUStARD dataset) descriptions achieved a higher CLIP similarity score in \textbf{78.42}\% of cases, compared to just \textbf{20.14}\% for the baseline descriptions. This quantitative evidence provides a compelling explanation for the downstream task improvements.

Additionally, we examined psychometric properties of the generated text. As summarized in Table \ref{tab:liwc-analysis} in Appendix, we found that preference tuning resulted in significant changes to the descriptions' average length and emotional tone, suggesting the VLM learns to produce stylistically different text that is more helpful for the agent. 

However, we also observed that the preference-tuned models were more prone to hallucination (e.g., describing a non-existent "voiceover"). It is not surprising since the language agent has no way to judge whether the explanation is faithful to the original video or not, which also aligns with data collection processes of RLHF. For example, in \cref{tab:generated-examples} found in the Appendix, we see that the VLM is describing a voiceover that does not exist. As a rough measure of hallucination, we count the occurrence of the word "voice" and found that the baseline model generates captions with 0 occurrence of "voice". However, the without GT, with GT, and with ICL models generate 5, 34, and 1 occurrences of this word. Despite the hallucinations, it is interesting that this resulted in a large performance improvement even when the visual descriptions were factually inaccurate. We will explore how to mitigate these sorts of hallucinations in the future.

\subsection{Human Evaluation of Preference Alignment}
To compare LLM agent's preferences to human preferences, we conducted a small scale human study. Participants were presented with randomly selected 12 video conversations, each accompanied by two distinct VLM-generated descriptions (detailed study is outlined in appendix \ref{human_strudy}). Out of 240 valid judgments, the human evaluators' preferences aligned with the LLM's preferences in 155 cases. This corresponds to a 64.6\% preference alignment rate. We then measured the preference alignment rate by comparing whether the human's preference matched the LLM agent's preference for the same pair. A closer, per-sample analysis reveals a consistent trend: for 9 out of the 12 video scenes, the majority of human participants' preferences aligned with the LLM agent's preference. For the remaining three scenes, one showed a majority preference that disagreed with the LLM, while the other two resulted in a tie in human judgments (\cref{tab:human_study_per_sample}). This result, indicates that the LLM's judgment in identifying helpful multimodal descriptions is robustly similar to human intuition.

\section{Discussion \& Conclusion}
Our results show that, the common approach of combining VLM-generated descriptions with text often yields negative results, where our proposed framework consistently solves this problem. The method proved effective across all six LLM agents on both sarcasm and humor datasets supporting our primary hypothesis: a unimodal language agent can provide effective preference feedback to align a multimodal model for its own downstream tasks.

A key insight from our analysis is that the observed improvement is not merely stylistic. The CLIP similarity analysis indicates that preference-tuned descriptions are substantially more aligned with the visual ground truth, suggesting that the LLM’s text-based preferences implicitly guide the VLM toward more accurate and contextually relevant descriptions. This, in turn, provides a clearer, less noisy signal to the agent, explaining the substantial improvements in downstream performance. Furthermore, the observed changes in linguistic properties, such as emotional tone, indicate that the VLM also learns how to structure these descriptions in a manner that is more conducive to the LLM's reasoning style.

However, the improvements have important nuances. For powerful models like Llama3.3-70b, the strong Utterance Only baseline sometimes remained superior, suggesting that the VLM descriptions do not always add sufficient value. This indicates that, larger models may already implicitly encode multimodal understanding, which can diminish the added value of external visual descriptions. We also observed occasional hallucinations in the preference-tuned outputs, where the model inferred non-existent visual details; an expected limitation of text-only feedback.  Our human study contextualizes these limitations, showing a strong but imperfect 64.6\% alignment between LLM and human preferences, which confirms the LLM is an effective but not flawless guide.

Overall, this work demonstrates that LLMs can effectively generate the preference signals needed to produce better textual descriptions for improving an LLM's multimodal understanding without costly retraining. Future work should focus on extending this successful feedback framework to more complex reasoning tasks and other modalities.

\section*{Limitation}
This study demonstrates the potential of Large Language Models (LLMs) to interpret multimodal contexts using text-only descriptions for multimodal social reasoning task, achieving promising results. Although our experiments demonstrate that offline reinforcement-learning (RL) fine-tuning can significantly improve policy performance on static datasets, this design lacks of real-time adaptation. In the current framework, DPO processes preference data in batch mode, optimizing the model based on a static dataset. This offline approach restricts the model’s ability to adapt dynamically to new or evolving data patterns, which is critical for real-world applications. To address this limitation, a primary focus can be on transitioning from offline to online RL fine-tuning. Online RL would enable the model to learn and adapt in real-time as new data becomes available, enhancing its responsiveness and accuracy in dynamic contexts. Furthermore, our experiments focus on social multimodal reasoning tasks such as sarcasm and humor detection, which depend on subtle contextual cues rather than low-level perception. As such, the framework may not directly generalize to perceptual tasks (e.g., counting or object recognition) where visual verification is essential. On the other hand, preference-tuned VLMs show clear improvements, but we occasionally observe hallucinated or exaggerated details in the generated descriptions. 

Future work should explore integrating vision-grounded rewards and hybrid alignment strategies to reduce hallucination, and extending this framework to more perception-heavy tasks (e,g. VQA) and additional modalities (such as audio).

\section*{Ethics Statement}
This study uses publicly available datasets (MUSTARD) and pre-trained models, following their respective licenses, and does not collect or process personal or sensitive data outside of the user study. The proxies for video data are generated from dataset content in accordance with academic use guidelines. In addition, this research included a human study. The study protocol was reviewed and approved by the Institutional Review Board. Participants were recruited through the Prolific academic crowd-sourcing platform. All participants provided informed consent prior to participation and were compensated at a rate of \$6 per half hour, based on the expected time for the task to which they were assigned. No personally identifiable information was collected, and all data from the study was anonymized to protect participant privacy. This research is intended for academic exploration and is not suitable for direct deployment in sensitive or decision-critical applications without further evaluation.

\section*{Acknowledgement}
Research was sponsored by the Army Research Laboratory and was accomplished under Cooperative Agreement Number W911NF-23-2-0224. The views and conclusions contained in this document are those of the authors and should not be interpreted as representing the official policies, either expressed or implied, of the Army Research Laboratory or the U.S. Government. The U.S. Government is authorized to reproduce and distribute reprints for Government purposes notwithstanding any copyright notation herein. We would also like to thank Anjila Budathoki for her invaluable assistance in setting up the user study.

\bibliography{reference}

\appendix

\section{Prompts to VLM for generating video descriptions}
\label{sec:vlm-prompts}

To ensure diverse description generation using the LLaVA-NeXT-Video model, we define five distinct prompts, each crafted to elicit different aspects of the video content. These prompts are:

\begin{lstlisting}
# Define 5 diverse prompts
diverse_prompts = [
    "Describe what is happening in this video in detail.",
    "Describe the video in such a way that it will be helpful for sarcasm detection. Try to keep the description brief.",
    "Describe the facial expressions and the speech tone of the speakers that is useful to understand the subtle meaning of the conversation. Try to keep the description brief.",
    "Describe the facial expressions in the video that might indicate contrasting emotions. Keep the description brief",
    "Provide a brief description of this video."
]
\end{lstlisting}

For each video, the model is run once per prompt, generating five separate outputs. No sampling strategies (such as varying temperature or top-$k$) are used; instead, diversity arises entirely from the distinct prompt formulations.

\section{Prompts for the Language Agent (Deepseek-R1)}
\subsection{Prompt to Rank the Captions without Ground Truth}
\label{prompt1}
\begin{lstlisting}
 prompt = f"""Consider that the visual description of a video scene 
 is: {video_description}.{raw_utterance_text}
        
Please rate the likelihood that the scene contains SARCASM on a scale
of 1-10.

1-5: NOT sarcastic
- 1: Strongly believe this is NOT sarcastic
- 5: Not sure, but leaning towards NOT sarcastic

6-10: Sarcastic
- 6: Not sure, but leaning towards sarcastic
- 10: Strongly believe this IS sarcastic

Please provide your response in this exact format:
FINAL SCORE: [number between 1-10]"""
\end{lstlisting}

\subsection{Prompt to Rank the Captions With Ground Truth}\label{prompt2}
\begin{lstlisting}
 prompt = f"""You are tasked with evaluating five video descriptions to determine how helpful each is in predicting whether an utterance is {sarcasm_status}. Below are the five video descriptions and the original utterance.

{raw_utterance_text}Video Descriptions:
1. {descriptions[0]}
2. {descriptions[1]}
3. {descriptions[2]}
4. {descriptions[3]}
5. {descriptions[4]}

Please rank each description based on how helpful it is for predicting that the utterance is {sarcasm_status}. Assign a score between 1 and 10, where:
- 1: Not helpful at all
- 10: Very helpful

Provide your response in this exact format:
Description 1: [score]
Description 2: [score]
Description 3: [score]
Description 4: [score]
Description 5: [score]
"""

\end{lstlisting}

\subsection{Prompt for Ranking Captions with Few-Shot Examples}\label{prompt5}
\begin{lstlisting}
  prompt = f"""You are given a video scene description, previous conversation context, and an utterance from that scene. Your job is to rate how likely the scene contains SARCASM, on a scale of 1 to 10, using this rubric:

1 to 5: NOT sarcastic
- 1: Strongly believe this is NOT sarcastic
- 5: Not sure, but leaning towards NOT sarcastic

6 to 10: Sarcastic
- 6: Not sure, but leaning towards sarcastic
- 10: Strongly believe this IS sarcastic

Always reply **exactly** in this format:
FINAL SCORE: [number between 1 to 10]
---
**Example 1**
Video description: In the video, we see a man and a woman in a hospital setting. The man, dressed in a white coat and tie, is holding a clipboard and appears to be a doctor. He is speaking to the woman, who is wearing a pink shirt, and seems to be explaining something to her. The woman is smiling and nodding along, indicating that she is engaged in the conversation. However, the man's tone is sarcastic and dismissive, as he talks about the woman's medical condition with a lack of concern and even makes a joke about it. The woman seems to be taking the situation lightly, laughing along with the man's jokes, which suggests that she is either in on the joke or is not bothered by his tone. The setting is a typical hospital environment, with medical equipment visible in the background, and the overall atmosphere is light-hearted and humorous.

Previous conversation:
RACHEL: "All right, I'm outta here!"
MONICA: "I'm kidding! I'm kidding!"
RACHEL: "So were done then!"
Utterance: PERSON: "Almost! But first, we gotta start."
FINAL SCORE: 10

**Example 2**
The video features two men sitting in a room, one of whom is holding a coffee cup and speaking to the other. The man holding the coffee cup is wearing a red and blue striped shirt, while the other man is wearing a white shirt. They are engaged in a conversation, and the man in the red and blue shirt is holding the coffee cup in his right hand. The room appears to be an office setting, with a desk and a chair visible in the background.

Previous conversation:
HOWARD: "Do you really think you should be drinking right now?"
RAJ: "How else am I supposed to talk to the Human Resources lady?"
HOWARD: "I don't know. Seek professional help?"
Utterance: RAJ: "I did. The guy at the liquor store said this stuff tastes great in coffee."
FINAL SCORE: 1
---
**Now you try**
Video description: {desc}
{context_text}{raw_utterance_text}
FINAL SCORE: [number between 1-10]
"""

\end{lstlisting}

\subsection{Prompt to Calculate Accuracy}\label{prompt4}
\begin{lstlisting}
prompt = f""" You are given a video scene description, previous conversation context, and an utterance from that scene. Your job is to rate how likely the scene contains SARCASM, on a scale of 1-10, using this rubric:

1-5: NOT sarcastic
- 1: Strongly believe this is NOT sarcastic
- 5: Not sure, but leaning towards NOT sarcastic

6-10: Sarcastic
- 6: Not sure, but leaning towards sarcastic
- 10: Strongly believe this IS sarcastic

Video description: {description}
Context: {context}
Target Utterance: {utterance}
"""
\end{lstlisting}
Depending on the evaluation condition, the \texttt{video\_description} comes from the output of the VLM. The \texttt{utterance} is provided using the text modality. Additionally, any conversation \texttt{context} is also provided.

\subsection{Zero-Shot VLM Evaluation}\label{vl-eval}

We provided the model with the video clip and the full dialogue context for each sample, then asked it to classify the final utterance. The full prompt template used for this evaluation is as below:

\begin{lstlisting}
    prompt_text = (

f"Here is the conversation from the video:{dialogue}"

Considering the full dialogue and the visual scene, is the final utterance by {task['speaker']} sarcastic? "

"Answer with only 'yes' or 'no'."

)
\end{lstlisting}
This baseline achieved an accuracy of 51.80\% on MUStARD, demonstrating that our preference-tuning method significantly outperforms the base VLM's standalone reasoning capabilities.
\subsection{Detailed Human Study on Preference Alignment} \label{human_strudy}
To validate that the preferences generated by our LLM agent we conducted a human evaluation study.

\subsubsection{Study Design}
We recruited total 20 participants through the Prolific academic crowd-sourcing platform. The study was designed and administered using Qualtrics survey software. After providing informed consent, participants were shown a detailed task description with examples explaining how to evaluate and score video scene descriptions .

The main task consisted of randomly selected 12 questions. For each question, participants were presented with the dialogue context and a target utterance from a video scene, along with two distinct textual descriptions of that scene (`Description A' and `Description B'). Their task was to score each description on a scale from 1 (strongly not sarcastic) to 10 (strongly sarcastic) based on how well it reflected the sarcasm in the target utterance. To ensure a clear preference was established between the two options, participants were explicitly instructed not to assign the same score to both descriptions. We used the following instruction for the participants.

\begin{lstlisting}
Task Description:
You will be shown pairs of descriptions generated for a video scene, along with the scene's context and a target utterance. Your task is to score the two descriptions based on how accurately it reflects the presence of sarcasm (if any) in the utterance, using this scale:

1-5: NOT sarcastic (1 = strongly not sarcastic, 5 = leaning not sarcastic)
6-10: Sarcastic (6 = leaning sarcastic, 10 = strongly sarcastic)

Explanation of Terms for Evaluators:
Sarcasm: Sarcasm's role in conversation is to allow individuals to say the opposite of what they actually mean, often to be funny or to mock something. 
Description: A text summary of the video scene, capturing visual and situational details. This gives you a better mental picture of the video scene and help you to decide whether the target utterance is sarcastic or not. 
Utterance: The specific spoken line from the scene, where sarcasm may be present.
Context: Background or prior dialogue in the scene that provides clues about sarcasm.

Additional Guidance:
1. If both descriptions suggest the utterance is sarcastic, score the better one higher (6-10).
2. If both suggest it's not sarcastic, score the better one lower (1-5).
3. Do not provide same score to both of the descriptions.
\end{lstlisting}

\subsubsection{Analysis and Result}
We analyzed the data collected from 20 participants who evaluated 12 pairs of video scene descriptions. To perform this analysis, for each of the 240 human judgments (20 participants × 12 samples), we determined the human's preference by comparing the 1-10 sarcasm score they assigned to Description `A' versus Description `B'. A higher score indicated a preference for that description in sarcastic scenes, while a lower score was preferred for non-sarcastic scenes. We then compared each human's preference with the LLM's preference for the same sample. This comparison yielded a ``match'' if the preferences were the same, a ``non-match'' if they were different.

The overall agreement between the human evaluators and the LLM agent was strong. Across all 240 judgments, the human preference matched the LLM's preference in 155 cases, and did not match in 85 cases. This corresponds to an overall preference alignment rate of 64.6\%, which is significantly higher than a 50\% random chance baseline (Z = 4.724, p = 0.000001). 

\begin{table}[ht]
\centering
\begin{tabular}{lccc}
\toprule
\textbf{Sample ID} & \textbf{Do Match} & \textbf{Do Not Match} \\
\midrule
1\_11006 & 10 & 10 \\
1\_11021 & 15 & 5 \\
1\_12202 & 12 & 8 \\
1\_3333  & 12 & 8 \\
1\_3837  & 15 & 5 \\
1\_507   & 16 & 4 \\
1\_8136  & 16 & 4 \\
1\_8417  & 15 & 5 \\
2\_147   & 6  & 14 \\
2\_279   & 17 & 3 \\
2\_380   & 10 & 10 \\
2\_49    & 11 & 9 \\
\midrule
\textbf{Total} & \textbf{155} & \textbf{85} \\
\bottomrule
\end{tabular}
\caption{Per-Sample Results of Human Study on Preference Alignment. This table presents the per-sample results of the human preference study, detailing the number of human judgments that aligned ("Do Match") or misaligned ("Do Not Match") with the LLM's preference for each of the 12 video scenes. The data shows a consistent trend, with a clear majority of human evaluators agreeing with the LLM's preference on 9 of the 12 samples.
}
\label{tab:human_study_per_sample}
\end{table}

We conducted a per-sample analysis, aggregating the judgments for each of the 12 video scenes individually. The results, shown in tab. \ref{tab:human_study_per_sample} , reveal a robust trend. For 9 out of the 12 samples, a clear majority of human participants' preferences aligned with the LLM's preference. Of the remaining three samples, only one showed a majority preference that disagreed with the LLM, and the other two resulted in a tie in human judgments. This strong overall alignment, combined with a consistent majority agreement on a per-sample basis, demonstrates that the LLM's judgment in identifying helpful multimodal descriptions is robustly similar to human intuition, validating its use as a source of preference feedback in our framework.
\subsection{Effect of beta} \label{beta}

\begin{table}[H]
  \centering
  \small
  \begin{tabular}{l|ccc}
    \toprule
    \textbf{($\beta$) Value} & \textbf{Accuracy \%} & \textbf{Precision} & \textbf{Recall} \\
    \midrule

     0.1 & \textbf{60.45} & \textbf{63.0} & \textbf{55.0} \\
    0.5 & 57.14 & 59.0 & 55.0 \\ 
    0.8 & 53.33   &    54.0   &    47.0      \\
    \bottomrule
  \end{tabular}
  \caption{Effect of ($\beta$) on DPO Fine-Tuning. At $\beta = 0.1$, we achieve best results. As we increase the value of $\beta$ we observe that accuracy precision, and recall becomes lower.}
  \label{tab:effect-of-different-beta-value}
\end{table}

\subsection{LIWC-22 Analysis}\label{liwc}
We conducted a psychometric analysis using LIWC-22 to understand what properties our preference-tuning framework imparts on the VLM-generated text. The result is shown in \cref{tab:liwc-analysis}. The result in indicates that the stylistic properties of the descriptions change significantly after preference tuning.
\begin{table*}[ht]
\centering \small
\begin{tabular}{l|rrr|rrr|rrr}
\toprule
            & \multicolumn{3}{c|}{\textbf{Overall}} & \multicolumn{3}{c|}{\textbf{Agent pos pred}}  & \multicolumn{3}{c}{\textbf{Agent neg pred}} \\ 
 \textbf{VLM Model}  
    & Acc \%  & Length        & Tone    
    & Acc \%   & Length      & Tone             
    & Acc \%   & Length       & Tone   \\ \midrule
Baseline          
    & 52.9   & 162.1         & 47.1
    & 58.0    & 148.0         & 41.4 
    & 50.0    & 170.3         & 50.3   \\ \midrule
\textbf{Preference Tuned}     &   &               &             &                    &   &                     &                  &     &                     \\ 
With GT       
    &  57.7     & 147.8         & 44.5 
    &  57.8     & 148.7         & 41.6              
    &  57.5     & 145.7         & 50.9                \\
    
Without GT     
    &   66.9    & 170.8         & 54.4      
    &   70.1    & 171.2         & 52.8  
    &   63.8    & 170.4         & 56.0                \\
With ICL 
    & 65.1      & 120.2         & 38.0     
    & 75.6      & 132.2         & 43.0                 
    & 59.5      & 113.8         & 35.2   \\ \bottomrule              
\end{tabular}
\caption{Summary statistics comparing text output from the VLM model for the MUStARD. Three conditions are compared: 1) overall statistics, 2) statistics of VLM outputs which led the agents to predict a positive label, and 3) summary of what caused the agent to predict a negative label. Tone scores are obtained from LIWC-22. After the VLM was preference tuned, the tone of the VLM also changed significantly. We can see that for the without GT and with ICL conditions, there was a large difference in tone for the tuned VLM outputs. We also found significant variability in the lengths of the generated text with ICL being the shortest. The VLM trained on preferences of the LLM when given the ground truth tended to have a shorter token length. In general the agent was more accurate with positive instances of sarcasm than negative ones.}
\label{tab:liwc-analysis}
\end{table*}

\subsection{Generated Output Examples}
\label{sec:gen-out}
To provide a qualitative understanding of how preference tuning alters the VLM's output, \cref{tab:generated-examples} presents several generated examples. The baseline model's description (Example 1) is typically factual and neutral, focusing on objects and settings (e.g., "a man is sitting on a couch, holding a phone"). In contrast, the preference-tuned models generate more interpretive and emotionally rich descriptions.

\begin{table*}[t]
\small
\begin{tabularx}{\linewidth}{X}
\toprule
\textbf{Agent provided preferences with knowledge of GT. After training, sometimes the VLM will hallucinate. For example, we see that the VLM is talking about a voiceover that doesn't exist.} \\
In the video, a man and a woman are seen lying on the ground, surrounded by fallen leaves, which suggests it's autumn. The man is wearing a blue sweater and the woman is in a blue top, both with their arms around each other. The man is speaking directly to the camera, and the woman is looking at it with a smile on her face. The man's tone is serious, while the woman's is more lighthearted. The man's voiceover explains that they're happy and in love, and the woman's voiceover is more playful, contrasting with his. The man's voiceover is in a serious tone, while the woman's is more lighthearted. The scene is set in a park with autumn leaves, and the man and woman are the only two people in it. The man's voiceover explains that they're happy and in love, while the woman's voiceover contrasts with his, adding a sense of humor to the scene. \\ \midrule
\textbf{In the following three examples, we see a different sentiment being expressed.} \\
\textbf{Example 1, baseline model:}
In the video, we see a man sitting on a couch, holding a phone to his ear. He appears to be engaged in a conversation, possibly a phone call or a video call, as he is looking at the camera with a focused expression. The room around him is casually furnished, with a couch, a chair, and some posters on the wall. The man's attire is a sweater with a pattern, and he seems to be in a relaxed setting, possibly at home. The lighting is warm, and the overall atmosphere is calm. The video captures a moment of everyday life, with the man's interaction with the phone being the main focus.\\
\textbf{Example 2, preference given with knowledge of GT:}
In the video, we see a man sitting on a couch and talking on a phone. He's wearing a sweater with a distinctive checkered pattern and appears to be in a casual setting, possibly his living room. His expression is serious, and he's engaged in a heated conversation. The camera zooms in on his face, emphasizing his intensity. The contrast between his calm exterior and the inner turmoil is evident.\\
\textbf{Example 3, preference given with in-context examples:}
In the video, we see a young man sitting comfortably on a couch in a living room. He is dressed in a cozy sweater and is engaged in a phone conversation. His face is lit up with a smile, suggesting a pleasant conversation. The room around him is warmly decorated, with a poster of a man on the wall and a bookshelf filled with various items. The man on the phone is holding a phone to his ear, indicating an ongoing call. The living room is well-lit, with a lamp casting a soft glow on the scene. The man's relaxed posture and the casual setting suggest a comfortable and familiar environment. The video captures a moment of everyday life, showcasing the man's engagement in a phone call while enjoying his surroundings.\\

\bottomrule
\end{tabularx}
\caption{Examples of VLM generated sequences.}
\label{tab:generated-examples}
\end{table*}

\subsection{Result Analysis for multiple agents and datasets}
\label{sec:other-results}

To assess the impact and generalizability of our proposed framework, we evaluated performance across six different agents on two distinct multimodal tasks: sarcasm detection (MUStARD) and humor detection (UR-FUNNY). The comprehensive results are presented in \cref{tab1}. 
\begin{table*}[t]
    \centering
    \small
    \begin{tabular}{l|l|ccc|ccc} 
        \toprule
         & & \multicolumn{3}{c|}{\textbf{MUStARD dataset}} & \multicolumn{3}{c}{\textbf{UR-FUNNY dataset}} \\ \cline{3-8}
        
        \textbf{Model} & \textbf{Condition} & \textbf{Accuracy} & \textbf{Precision} & \textbf{Recall} & \textbf{Accuracy} & \textbf{Precision} & \textbf{Recall} \\
        \midrule

        \multirow{7}{*}{Deepseek-r1 -7b} & \textbf{Baseline} & & & & & & \\
         & { } Utterance Only & 61.2 & 62.0 & 64.0 & \textbf{57.8} & 62.0 & 64.0 \\
         & { } Visual Only  & 48.9 &  52.0 & 19.0 & 51.6 & 53.0 & 19.0  \\
         & { } Multimodal & 52.9 & 58.0 & 40.0 & 55.6 & 55.0 & 64.0 \\ \cline{2-8}
         &\textbf{Preference Tuned }& & & & & & \\
         & { } Visual Only & 50.36  & 71.0  & 7.0 & 51.6 & 53.0 & 16.0  \\
         & { } Multimodal & \textbf{66.9} & 70.0 & 65.0 & 56.5 & 56.0 & 60.0 \\
        \midrule

        \multirow{7}{*}{Qwen-7b} & \textbf{Baseline} & & & & & & \\
         & { } Utterance Only & 51.8 & 52.0 & 97.0 & 48.3 & 49.0 & 94.0 \\
         & { } Visual Only & 55.4 & 54.0 & 10.0 & 49.3 & 49.0 & 99.0 \\
         & { } Multimodal & 53.3 & 55.0 & 78.0 & 47.4 & 48.0 & 89.0 \\ \cline{2-8}
         &\textbf{Preference Tuned }& & & & & & \\
         & { } Visual Only & 51.8 & 52.0 & 10.0 & 49.8 & 50.0 & 92.0 \\
         & { } Multimodal & \textbf{61.2} & 62.0 & 83.0 & \textbf{49.0} & 49.0 & 98.0 \\
        \midrule

        \multirow{7}{*}{Vicuna-30b} & \textbf{Baseline} & & & & & & \\
         & { } Utterance Only & \textbf{56.1} & 54.0 & 99.0 & 49.9 & 50.0 & 99.0 \\
         & { } Visual Only & 51.7 & 53.0 & 93.0 & 49.5 & 49.0 & 87.0 \\
         & { } Multimodal & 48.3 & 50.0 & 90.0 & 49.4 & 49.0 & 10.0 \\ \cline{2-8}
         &\textbf{Preference Tuned }& & & & & & \\
         & { } Visual Only & 49.6 & 51.0 & 49.0 & 51.1 & 50.0 & 84.0 \\
         & { } Multimodal & 52.7 & 70.0 & 65.0 & \textbf{50.7} & 70.0 & 65.0 \\
        \midrule

        \multirow{7}{*}{Lama2-7b} & \textbf{Baseline} & & & & & & \\
         & { } Utterance Only & 53.5 & 53.0 & 74.0 & 49.5 & 49.0 & 10.0 \\
         & { } Visual Only & 54.0 & 53.0 & 99.0 & 49.3 & 49.0 & 99.0 \\
         & { } Multimodal & 50.8 & 58.0 & 40.0 & 49.3 & 49.0 & 10.0 \\ \cline{2-8}
         &\textbf{Preference Tuned }& & & & & & \\
         & { } Visual Only & 54.0 & 53.0 & 96.0 & 49.0 & 49.0 & 99.0 \\
         & { } Multimodal & \textbf{56.5} & 58.0 & 42.0 & \textbf{53.5} & 53.0 & 74.0 \\
        \midrule

        \multirow{7}{*}{Llama3.3-70b} & \textbf{Baseline} & & & & & & \\
         & { } Utterance Only & \textbf{71.9} & 65.0 & 99.0 & \textbf{73.0} & 68.0 & 85.0 \\
         & { } Visual Only & 52.5 & 72.0 & 14.0 & 50.6 & 50.0 & 20.0 \\
         & { } Multimodal & 64.0 & 59.0 & 99.0 & 67.8 & 62.0 & 92.0 \\ \cline{2-8}
         &\textbf{Preference Tuned }& & & & & & \\
         & { } Visual Only & 47.5 & 47.0 & 10.0 & 51.9 & 54.0 & 17.0 \\
         & { } Multimodal & 66.2 & 61.0 & 10.0 & 69.6 & 63.0 & 91.0 \\
        \midrule

        \multirow{7}{*}{Gemma2-2b} & \textbf{Baseline} & & & & & & \\
         & { } Utterance Only & 52.5 & 52.0 & 10.0 & 50.1 & 50.0 & 98.0 \\
         & { } Visual Only & 54.0 & 55.0 & 58.0 & 52.0 & 52.0 & 43.0 \\
         & { } Multimodal & 53.7 & 54.0 & 96.0 & 50.5 & 50.0 & 95.0 \\ \cline{2-8}
         &\textbf{Preference Tuned }& & & & & & \\
         & { } Visual Only & 51.8 & 73.0 & 11.0 & 49.3 & 49.0 & 79.0 \\
         & { } Multimodal & \textbf{56.3} & 56.0 & 88.0 & \textbf{57.7} & 56.0 & 62.0 \\
        \bottomrule
    \end{tabular}
    \caption{Performance metrics with different agents. The table demonstrates the effectiveness and generalizability of the proposed method for fine-tuning a VLM based on preference feedback from a unimodal LLM. The experiments are conducted across two distinct datasets, MUStARD for sarcasm detection and UR Funny for humor detection and evaluated using six different LLM agents. The result shows that fine-tuning a VLM using preference feedback from a LLM agent consistently improves performance on multimodal tasks while using descriptions from an un-tuned VLM degrades the accuracy of an LLM agent on both the MUSTARD and UR Funny datasets. For many agents, the fine-tuned descriptions help surpass the text-only accuracy. In other cases, the "utterance only" performance remains the highest, even though the fine-tuned description is still an improvement over the un-tuned one.}
    \label{tab1}
\end{table*}

\end{document}